\pdfoutput=1

\documentclass[11pt]{article}

\usepackage[final]{acl}

\usepackage{times}
\usepackage{latexsym}
\usepackage{MnSymbol}
\usepackage{amsmath}
\usepackage[normalem]{ulem}
\usepackage[T1]{fontenc}

\usepackage[utf8]{inputenc}

\usepackage{microtype}

\usepackage{inconsolata}

\usepackage{graphicx}
\usepackage{enumitem}
\usepackage{cleveref}

\title{
    CutPaste\&Find: Efficient Multimodal Hallucination Detector with Visual-aid Knowledge Base
}

\author{
    Cong-Duy Nguyen\textsuperscript{\rm 1} \quad
    Xiaobao Wu\textsuperscript{\rm 1}\thanks{Corresponding Authors.} \quad
    Duc Anh Vu\textsuperscript{\rm 1} \\ 
    \textbf{Shuai Zhao}\textsuperscript{\rm 1} \quad 
    \textbf{Thong Nguyen}\textsuperscript{\rm 2} \quad 
    \textbf{Anh Tuan Luu}\textsuperscript{\rm 1}$^*$ \\
  $^1$Nanyang Technological University, Singapore \\
  $^2$National University of Singapore, Singapore\\ 
  \texttt{ \{nguyentr003,xiaobao.wu,vuducanh001,shuai.zhao,anhtuan.luu\}@ntu.edu.sg } \\
    \texttt{thongnguyen050999@gmail.com}
  }

\begin{document}
\maketitle
\begin{abstract}
Large Vision-Language Models (LVLMs) have demonstrated impressive multimodal reasoning capabilities, but they remain susceptible to hallucination, particularly object hallucination where non-existent objects or incorrect attributes are fabricated in generated descriptions. Existing detection methods achieve strong performance but rely heavily on expensive API calls and iterative LVLM-based validation, making them impractical for large-scale or offline use. To address these limitations, we propose CutPaste\&Find, a lightweight and training-free framework for detecting hallucinations in LVLM-generated outputs. Our approach leverages off-the-shelf visual and linguistic modules to perform multi-step verification efficiently without requiring LVLM inference. At the core of our framework is a Visual-aid Knowledge Base that encodes rich entity-attribute relationships and associated image representations. We introduce a scaling factor to refine similarity scores, mitigating the issue of suboptimal alignment values even for ground-truth image-text pairs. Comprehensive evaluations on benchmark datasets, including POPE and R-Bench, demonstrate that CutPaste\&Find achieves competitive hallucination detection performance while being significantly more efficient and cost-effective than previous methods.
\end{abstract}

\section{Introduction}


Large Vision Language Models (LVLMs) \cite{liu2023llava, ye2023mplug, zhu2023minigpt, li2023otter, bai2023qwen,dai2023instructblip} have exhibited impressive multimodal understanding and reasoning abilities.
They excel at various vision-language tasks, such as image captioning, visual question answering (VQA), and image-based dialogue, where they generate descriptive text based on visual inputs.
For example, models such as LLaVA~\cite{liu2023llava} and MiniGPT-4~\cite{zhu2023minigpt} can generate detailed captions from an image, answering complex questions about visual content.
However, LVLMs face one critical challenge: object hallucination.
This means LVLMs generate inconsistent or fabricated descriptions of a given image, such as inventing non-existent objects.
This challenge undermines the reliability and accuracy of LVLMs, severely limiting their broader applications.
To address this challenge, some hallucination detection methods have been proposed.
For example, Woodpecker~\cite{yin2023woodpecker}  is a training-free post-processing method that detects and corrects hallucinations in LVLMs.
LogicCheckGPT~\cite{wu2024logical} is a plug-and-play framework that detects and mitigates object hallucinations in LVLMs by probing their logical consistency in responses.

\begin{table}[t]
\centering
\resizebox{1.0 \columnwidth}{!}{
\begin{tabular}{l|c|c}
& \begin{tabular}[c]{@{}c@{}}Number of time\\ calling OpenAI api\\ for 1 sentence\end{tabular} & Module usage                                                           
\\ \hline
Woodpecker    & 4                                                                                            & \begin{tabular}[c]{@{}c@{}}GroundDINO, \\ BLIP2, QA2Claim\end{tabular}  \\ \hline
LogicCheckGPT & 6                                                                                            & LVLM (MPlug)                                                            \\ \hline
Our           & 0                                                                                            & \begin{tabular}[c]{@{}c@{}}SGP,  GroundDINO,\\ BLIP2, CLIP\end{tabular}
\end{tabular}}
\caption{Statistics of Calling OpenAI API and Module usage of LVLM hallucination detections.}
\label{tab:stat}
\vspace{-10pt}

\end{table}

"However, these methods have several limitations. First, they rely heavily on external API calls.
As reported in \Cref{tab:stat}, Woodpecker and LogicCheckGPT make 4-6 API calls to detect each sentence.
This dependence also limits offline usability, making deployment infeasible in scenarios with restricted API access or privacy concerns.
Second, they require iterative interactions with LVLMs to detect hallucinations.
For instance, LogicCheckGPT generates multiple queries and performs consistency checks across responses. This iterative process not only increases latency but also amplifies token consumption, making it impractical for large-scale applications or real-time inference.




To resolve these limitations, we propose a novel hallucination detection framework, CutPaste\&Find. 
It is training-free, offline, and computationally efficient.
Previous detection methods rely on external APIs. 
Differently, we validate visual entities with the pipeline by utilizing off-the-shelf textual and visual modules which are less hallucinated than LVLM~\cite{Yin_2024}.
Moreover, we build a visual knowledge base to improve the reliability without high-computational interaction.

While these methods achieve strong performance, they are resource-intensive and costly due to frequent API calls and the computational overhead of LVLM inference. Moreover, they are not suitable for offline use. To address these limitations, we propose a novel framework called CutPaste\&Find that is training-free, and efficient. Our framework leverages off-the-shelf modules to preprocess and perform multi-step inference without relying on powerful APIs. 

Additionally, to further enhance our framework, we introduce a Visual-aid Knowledge Base as a robust solution for detecting multimodal hallucinations. Unlike traditional methods, our approach is grounded in a curated knowledge base built on human-annotated data, ensuring high accuracy and reliability. Specifically, this knowledge base is constructed through process called Cut-and-Paste using the Visual Genome dataset~\cite{krishna2016visualgenomeconnectinglanguage}, which incorporates rich information, including object names, attributes, relationships, and corresponding images.

The proposed method compares the visual representation similarity score of a cropped image (or the entire image) with entity-attribute pairs or triplet relations retrieved from the knowledge base. To improve the accuracy of alignment interpretation, we introduce a scaling factor with visual-aid prior knowledge. This innovation addresses the observation that similarity scores often deviate from ideal values near 1.0, even for ground-truth pairs. By incorporating this scaling factor, our framework refines similarity scores, enhancing their ability to represent true alignment between model outputs and the underlying image.

In summary, our main contributions are as follows:

\begin{itemize}[leftmargin=*,itemsep=0pt]

\item  We propose a training-free, lightweight, and efficient framework named ABC for detecting hallucinations in LVLMs. This framework employs off-the-shelf modules to perform multi-step detection with minimal resources and without reliance on costly APIs.

\item We introduce a clean and robust Visual-aid Knowledge Base to enhance hallucination validation. This knowledge base integrates prior knowledge to improve detection accuracy.

\item We evaluate the effectiveness of our framework through extensive experiments on benchmark datasets, including POPE~\cite{li2023evaluating} (with the split from \cite{wu2024logical}) and R-Bench~\cite{pmlr-v235-wu24l}. Our results demonstrate the robustness and efficiency of the proposed approach.

\end{itemize}

\section{Related Work}

\subsection{LVLMs}

With the rapid advancement of large language models (LLMs) \cite{ouyang2022training, zhao2023survey,pan2024fallacy,wu2024akew,wu2024antileak,wu2024fastopic,wu2024survey},
integrating their general intelligence into multimodal domains has garnered significant interest. This has led to the emergence of large vision-language models (LVLMs) \cite{ye2023mplug, zhu2023minigpt, liu2023llava, li2023otter, dai2305instructblip, bai2023qwen}, designed to understand and generate multimodal content under instructions. Most LVLMs adopt a two-stage training paradigm: multimodal alignment pre-training followed by instruction tuning, where an alignment module processes multimodal inputs before passing them to an LLM for response generation. For instance, mPLUG-Owl \cite{ye2023mplug} pre-trains both the encoder and alignment module before fine-tuning LLaMa \cite{touvron2023llama} with low-rank adaptation. LLaVA \cite{liu2023llava} pre-trains the alignment network and fine-tunes it alongside Vicuna \cite{chiang2023vicuna} on constructed instructions. In contrast, MiniGPT-4 \cite{zhu2023minigpt} only fine-tunes the cross-modal alignment network while keeping other components frozen.
\subsection{Hallucination in LVLMs}

\begin{figure*}[t]

    \centering
    \includegraphics[width=0.99\textwidth]{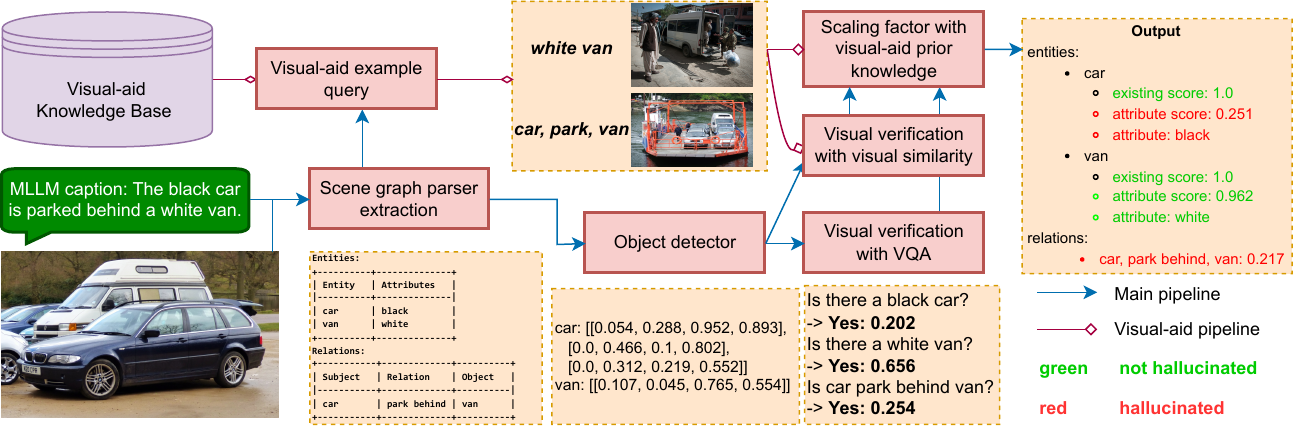}
    \caption{The overall architecture of CutPaste\&Find. The system extracts scene graphs from captions, verifies object existence and attributes using a knowledge base, and employs object detection, visual similarity, and VQA for validation. A scaling factor adjusts similarity scores, enhancing reliability, with hallucinated entities and relations highlighted in red. }
    \label{fig:model}
   \vspace{-10pt}
\end{figure*}

Despite their strong capabilities, LVLMs often suffer from hallucination issues. To address this, several benchmarks \cite{fu2023mme, xu2023lvlm, Li-hallucination-2023, lovenia2023negative, jing2023faithscore, chen2024unified} have been proposed to evaluate hallucination severity in LVLMs. 
Existing mitigation strategies can be categorized into three groups. The first and most common approach \cite{liu2023mitigating, gunjal2023detecting, lee2023volcano, wang2023vigc} relies on instruction tuning and retraining. LRV-Instruction \cite{liu2023mitigating} constructs a diverse dataset with both positive and negative instructions, while \cite{wang2023vigc} employs iterative instruction generation to enhance diversity and accuracy. Volcano \cite{lee2023volcano} enables self-feedback training for response revision. However, these methods depend heavily on high-quality instruction datasets and require substantial computation.
The second group focuses on decoding strategies to mitigate hallucinations. OPERA \cite{huang2023opera} introduces a penalty-based decoding method with a rollback strategy to reduce overconfidence, while VCD \cite{leng2023mitigating} applies contrastive decoding to minimize reliance on spurious biases. However, accessing LVLMs' internal states during decoding remains challenging for general users.
The third category integrates external models to refine LVLM outputs. Woodpecker~\cite{yin2023woodpecker} and LURE~\cite{zhou2023analyzing} leverage external detectors or specialized LVLMs as revisors to improve visual understanding. 
However, these methods depend on auxiliary models rather than enhancing the base LVLM itself. 
LogicCheckGPT~\cite{wu2024logical} takes a different approach by probing logical consistency in object-attribute relationships to detect and mitigate object hallucinations.

Complementary to these methods, a growing body of research aims to reduce hallucination at its semantic representation level. Works from \citet{nguyen2021enriching} and \citet{nguyen2024encoding} demonstrate that explicit global semantics control helps anchor generations to context. In a similar vein, \citet{nguyen2021contrastive, nguyen2024topic, nguyen2024topic_aware} show that refining latent topic structures can improve factual consistency in text generation.

In multimodal settings, angular margin-based contrastive learning \citep{nguyen2024kdmcse, nguyen2022adaptive} has been effective in tightening inter-modal alignment, which is crucial for grounding. Likewise, models like \citep{vo2024ti} and \citep{nguyen2023expand} enhance grounding via stronger joint embeddings or partial alignment. These representation-level improvements are vital for hallucination prevention, especially when training or decoding resources are limited.

Hallucination is further exacerbated in video-language models, where temporal reasoning adds complexity. Models like DemaFormer \citep{nguyen2023demaformer}, READ \citep{nguyen2024read} introduce novel architectures or adapters to better model temporal grounding. Motion-aware and multi-scale contrastive learning further improve alignment between textual queries and dynamic visual content—reducing hallucinations by anchoring generations in fine-grained temporal cues.

Finally, advances in neural topic modeling (e.g., \citep{wu2024fastopic, wu2023infoctm, wu2024survey, wu2024affinity, wu2024modeling, wu2023effective} contribute insights into how interpretable, structured representations can guide generation models. This direction suggests that structured semantic scaffolding—via topic modeling or contrastive optimization—may offer long-term solutions to hallucination by grounding language in disentangled conceptual spaces.

\section{Method}


In this section, we introduce the framework, first including Scene graph parser extraction, Object detector, Visual verification and then go through Scaling factor with visual-aid prior knowledge. We visualize the overall architecture in Figure~\ref{fig:model}.



\subsection{Scene graph parser extraction}
\label{sec:method_1}




Captions often emphasize key concepts, and the first step in our framework involves extracting these concepts from the generated sentences. Unlike prior works that primarily focus on the main objects mentioned in a sentence, we extract entities alongside their attributes and the relationships between these entities. For example, given the sentence, “\texttt{The man is wearing a black hat},” we extract entities (\texttt{man}, \texttt{hat}) and their attributes (\texttt{black}) and relationships (\texttt{wearing}). These serve as the foundation for subsequent hallucination diagnosis.

To achieve this, we leverage a Scene graph parser module (SGP), which efficiently converts unstructured textual descriptions into structured scene graphs. While large language models (LLMs) possess strong generalization capabilities and world knowledge, relying on them for this relatively straightforward task is resource-intensive and costly, particularly when APIs or open-source LLMs are used. SGP provides a more efficient alternative. Given a sentence $s$, SGP outputs two lists (a visualization is shown under Scene graph parser extraction in Figure~\ref{fig:model}):

\begin{align}
    E &= [ (e_1,a_1), (e_2,a_2), ... ] \\
    R &= [ (s_1,r-1,o_1), (s_2,r-2,o_2), ... ]
\end{align}

\noindent $e_i$ represents an entity and $a_i$ represents its attribute (if applicable), $(s_i,r_i,o_i)$ represents a relationship triplet of subject, relation, and object.

\subsection{Object detector}
\label{sec:method_2}

Using the entities extracted by the SGP, we apply an open-set object detector (OD) to locate each entity in the image. For a given entity $e_i$, if OD successfully detects it, we obtain a list of bounding boxes $B_i = \{b_{ia}, b_{ib}, ... \} $, where each bounding box $b_{ij}$  includes the top-left position $(x, y)$ and dimensions $(h, w)$. If $e_i$ cannot be detected, it is flagged as potentially hallucinated and validated using only the Visual Question Answering (VQA) module in the next step.

\subsection{Visual verification}


This stage validates the existence of entities, attributes, and relationships using two complementary methods: Visual Question Answering (VQA) and Visual Similarity.



\subsubsection{Visual verification with VQA}

We utilize a pre-trained VQA model~\cite{li2023blip}  to answer context-specific questions about the image. Compared to LVLMs, the VQA model produces concise responses with fewer hallucinations, making it a suitable choice. We formulate three types of questions:

\paragraph{Attribute asking} Given the entity $e_i$ and attribute $a_i$, we ask the VQA model ``\texttt{Is the $\{e_i\} \{a_i\}$?}''.

\paragraph{Relation asking} Given the triplet $s_i,r_i,o_i$, we ask the VQA model ``\texttt{Is the $\{s_i\} \{r_i\} \{o_i\}$?}''.

\paragraph{Existance asking} This format is used when the entity $e_i$ was missed from previous step~\ref{sec:method_2}. Given the entity $e_i$, we ask the VQA model ``\texttt{Is there \{$s_i$\} in the image}''.

We ask the VQA model in a binary Yes/No question, and instead of using the generated answer (``\texttt{yes}'' or ``\texttt{no}''), we base it on the scoring of ``\texttt{yes}'' token of the predicted probabilities of the output. Thus we can have a soft-predicted score $s^{qa}_i$ instead of a hard prediction.
Instead of relying on the binary outputs (\texttt{yes} or \texttt{no}), we use the probability of the \texttt{yes} token to derive a soft-predicted score $s^{qa}_i$, offering a more nuanced assessment.

\subsubsection{Visual verification with Visual Similarity}


To complement VQA, we validate hallucination using Visual-aid Knowledge Base. For an entity-attribute pair ($a_i$,$e_i$), we retrieve a list of representative images $\hat{C}_i$ from the datastore and compute visual representations $f_i$ (for the cropped image) and $\hat{F}_i$ (for retrieved images) using a pre-trained visual encoder (e.g., CLIP). The similarity score $s^{v}_i$ is calculated between $f_i$ and $\hat{F}_i$; lower scores indicate higher likelihoods of hallucination. A similar approach is applied to validate relationships $s_i,r_i,o_i$.

\subsection{Scaling factor with Visual-aid prior knowledge}



Similarity scores often deviate from ideal values near 1.0, even for ground-truth pairs. To address this, we introduce a scaling factor derived from the prior knowledge embedded in the datastore. This adjustment aligns similarity scores with realistic expectations, improving the sensitivity and robustness of hallucination detection.

\begin{figure}[t]

    \centering
    \includegraphics[width=0.45\textwidth]{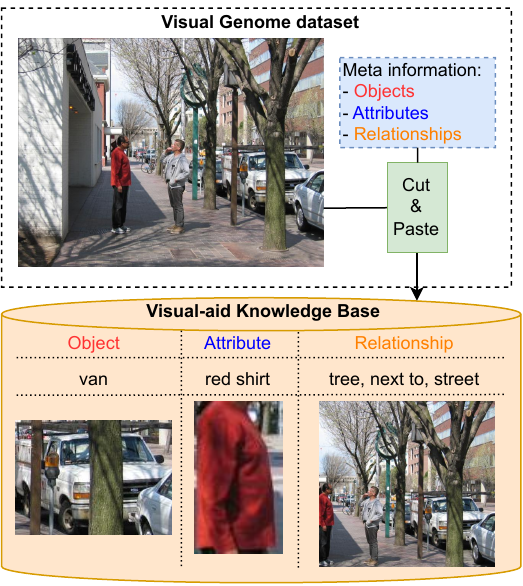}
    \caption{The figure illustrates a Visual-aid Knowledge Base built from meta-information and images, including objects, attributes, and relationships. A "Cut \& Paste" process segments visual elements such as objects (e.g., van), attributes (e.g., red shirt), and relationships (e.g., tree next to street) into a structured knowledge base. }
    \label{fig:kb}
    \vspace{-10pt}
\end{figure}

\paragraph{Visual Scaling Factor}
The scaling factor for visual encoding is computed as:

\vspace{-5pt}
\begin{align}
    &\tilde{S}^v_i = f(\hat{F}_i, \hat{F}_i^\intercal) \\
    &diag(\tilde{S}^v_i) = 0 \\
    &d^v_i = \max(\tilde{S}^v_i)
\vspace{-5pt}
\end{align}

\noindent where $d^v_i$ is the scaling factor for the visual encoding method. Then we calculate re-scale similarity score of an entity $e_i$:

\vspace{-5pt}
\begin{align}
    \bar{s}^{v}_i = s^{v}_i / d^v_i
\vspace{-5pt}
\end{align}

\paragraph{VQA Scaling Factor}
For the VQA module, we compute the scaling factor:

\vspace{-5pt}
\begin{align}
    & \tilde{S}^{qa}_i = [\tilde{s}^{qa}_{i1}, \tilde{s}^{qa}_{i2}, ...\tilde{s}^{qa}_{ij} ] \\
    & diag(\tilde{S}^qa_i) =0 \\
    & d^{qa}_i = max(\tilde{S}^{qa}_i)
\vspace{-5pt}
\end{align}

\noindent where $d^{qa}_i$ is the scaling factor for the visual question module. Then we calculate the re-scale similarity score of an entity $e_i$:

\vspace{-5pt}
\begin{equation}   
    \bar{s}^{qa}_i = s^{qa}_i / d^{qa}_i
\vspace{-5pt}
\end{equation}

\paragraph{Final output}


Given a generated caption $s$ and an image $I$, our classifier outputs a hallucination dictionary:

\begin{itemize}
    
\item Entity: $\left [ e_1: \tilde{s}_{e1}, \cdots, e_i: \tilde{s}_{ei} \right ]$

\item Attribute-entity: $\left [ e_1,a_1: \tilde{s}_{a1}, \cdots, e_i,a_i: \tilde{s}_{ai} \right ]$

\item Relation: $\left [ s_1,r_1,o_1: \tilde{s}_{r1}, \cdots, s_j,r_j,o_j: \tilde{s}_{ri} \right ]$

\end{itemize}

\noindent where $s_e, s_a, s_r$ are the existing, attribute, and relation score respectively. This structured output enables precise identification and correction of hallucinations, ensuring multimodal consistency.


\section{Visual-aid Knowledge Base}

\subsection{Cut-And-Paste}


Our Visual-aid Knowledge Base (VaKB) is built upon the Visual Genome dataset, a rich resource for tasks requiring fine-grained visual understanding. The VG includes 101,174 images sourced from MSCOCO, accompanied by 1.7 million question-answer pairs, averaging 17 questions per image. Unlike the traditional VQA datasets, Visual Genome offers a balanced distribution of six question types: What, Where, When, Who, Why, and How, making it particularly suitable for diverse multimodal reasoning tasks. Furthermore, it provides 108,000 images densely annotated with objects, attributes, and relationships, making it an invaluable resource for constructing comprehensive knowledge bases that bridge visual and textual modalities. We will pre-process the dataset with its annotation through process call "Cut-And-Paste".

\begin{table}[t]

\begin{center}
\scalebox{0.83
}{
\begin{tabular}{lcccc}
\hline
\multicolumn{1}{l|}{Method}     & \multicolumn{1}{c|}{Acc}   & \multicolumn{1}{c|}{Precision} & \multicolumn{1}{c|}{Recall} & F1 Score \\ \hline

\multicolumn{5}{c}{\textit{Adversarial}  }                                                                                                     \\ \hline
\multicolumn{1}{l|}{Woodpecker$^\diamondsuit$} & \multicolumn{1}{c|}{90.67} & \multicolumn{1}{c|}{90.13}     & \multicolumn{1}{c|}{91.33}  & 90.73    \\ \hline
\multicolumn{1}{l|}{LC-GPT$^\diamondsuit$}      & \multicolumn{1}{c|}{83.33} & \multicolumn{1}{c|}{\textbf{94.64}}     & \multicolumn{1}{c|}{70.67}  & 80.92    \\ \hline
\multicolumn{1}{l|}{CutPaste\&Find}        & \multicolumn{1}{c|}{\textbf{94.67}} & \multicolumn{1}{c|}{92.95}     & \multicolumn{1}{c|}{\textbf{96.67}}  & \textbf{94.77}    \\ \hline

\multicolumn{5}{c}{\textit{Popular}   }                                                                                                        \\ \hline
\multicolumn{1}{l|}{Woodpecker$^\diamondsuit$} & \multicolumn{1}{c|}{89.67} & \multicolumn{1}{c|}{91.03}     & \multicolumn{1}{c|}{88.00}  & 89.49    \\ \hline
\multicolumn{1}{l|}{LC-GPT$^\diamondsuit$}      & \multicolumn{1}{c|}{85.00} & \multicolumn{1}{c|}{\textbf{94.87}}     & \multicolumn{1}{c|}{74.00}  & 83.15    \\ \hline
\multicolumn{1}{l|}{CutPaste\&Find}        & \multicolumn{1}{c|}{\textbf{93.67}} & \multicolumn{1}{c|}{93.96}     & \multicolumn{1}{c|}{\textbf{93.33}}  & \textbf{93.65}    \\ \hline

\multicolumn{5}{c}{\textit{Random}     }                                                                                                       \\ \hline
\multicolumn{1}{l|}{Woodpecker$^\diamondsuit$} & \multicolumn{1}{c|}{93.33} & \multicolumn{1}{c|}{94.52}     & \multicolumn{1}{c|}{92.00}  & 93.24    \\ \hline
\multicolumn{1}{l|}{LC-GPT$^\diamondsuit$}      & \multicolumn{1}{c|}{85.67} & \multicolumn{1}{c|}{\textbf{98.20}}     & \multicolumn{1}{c|}{72.67}  & 83.52    \\ \hline
\multicolumn{1}{l|}{CutPaste\&Find}        & \multicolumn{1}{c|}{\textbf{95.67}} & \multicolumn{1}{c|}{95.36}     & \multicolumn{1}{c|}{\textbf{96.00}}  & \textbf{95.68}    \\ \hline

\end{tabular}
}

\caption{Results on POPE. The best performances within each setting are \textbf{bolded}. $\diamondsuit$ We run these models using the OpenAI GPT-4o version. Instead of passing non-existent claims as done in their implementation, we convert all yes-no questions into affirmative statements.}

\label{tab:main}
\end{center}

\vspace{-10pt}
\end{table}

\subsection{Datastore}



Each image $I$ in the Visual Genome dataset is enriched with human annotations, including entities, attributes, and relationships. These annotations are structured as follows:

\vspace{-5pt}
\paragraph{Entities} Each entity includes the object name and its bounding box in  $(x, y, w, h)$ format.
\vspace{-5pt}
\paragraph{Attributes} Each entity is further extended with a list of associated attributes, providing additional descriptive detail.
\vspace{-5pt}
\paragraph{Relationships} Each relationship annotation is represented as a triplet $(s,r,o)$, where $s$ is the subject, $r$ is the relationship, and $o$ is the object.

To build the VaKb, we crop the bounding boxes of entities and store both the cropped and original images in a database. This ensures that for each key $k$, we have a list of cropped images and their corresponding full images. Keys are constructed from entities, attribute-entity pairs, and relationship triplets.
Formally, we define the datastore as a collection of key-value pairs $(k_i, v_i)$, where the key $k_i$ can represent an entity $e_i$, an attribute-entity pair $(e_i, a_i)$, or a relationship triplet $(s_i, r_i, o_i)$. The value $v_i$ consists of the associated cropped and original images. To facilitate retrieval, each key $k_i$ is mapped to a fixed-length vector representation $f^T(k_i)$, computed by a pre-trained text encoder (e.g., CLIP-Text).
The datastore $\left( \mathcal{K}, \mathcal{V} \right)$ is constructed as:

\vspace{-10pt}
\begin{align}
\left( \mathcal{K}, \mathcal{V} \right) = \{ (f^T(k_i), v_i) | (k_i, v_i) \in \mathcal{D}, \\ k_i \in \{ e_i,(e_i,a_i),(s_i,r_i,o_i) \} \}
\vspace{-10pt}
\end{align}

Here, $\mathcal{D}$ represents the annotated examples in the Visual Genome dataset.

\begin{table}[t]

\begin{center}
\scalebox{0.83}{
\begin{tabular}{lcccc}
\hline
\multicolumn{1}{l|}{Method}     & \multicolumn{1}{c|}{Acc}   & \multicolumn{1}{c|}{Precision} & \multicolumn{1}{c|}{Recall} & F1 Score \\ \hline

\multicolumn{5}{c}{\textit{Image level}    }                                                                                                         \\ \hline
\multicolumn{1}{l|}{Woodpecker$^\diamondsuit$} & \multicolumn{1}{c|}{55.00} & \multicolumn{1}{c|}{55.03}     & \multicolumn{1}{c|}{54.67}  & 54.85    \\ \hline
\multicolumn{1}{l|}{LC-GPT$^\diamondsuit$}      & \multicolumn{1}{c|}{62.00} & \multicolumn{1}{c|}{66.07}     & \multicolumn{1}{c|}{49.33}  & 56.49    \\ \hline
\multicolumn{1}{l|}{CutPaste\&Find}        & \multicolumn{1}{c|}{\textbf{79.00}} & \multicolumn{1}{c|}{\textbf{74.58}}     & \multicolumn{1}{c|}{\textbf{88.00}}  & \textbf{80.73}    \\ \hline
\multicolumn{5}{c}{\textit{Instance level}   }                                                                                                \\ \hline
\multicolumn{1}{l|}{Woodpecker$^\diamondsuit$} & \multicolumn{1}{c|}{56.67} & \multicolumn{1}{c|}{57.04}     & \multicolumn{1}{c|}{54.00}   & 55.48    \\ \hline
\multicolumn{1}{l|}{LC-GPT$^\diamondsuit$}      & \multicolumn{1}{c|}{60.67} & \multicolumn{1}{c|}{62.12}     & \multicolumn{1}{c|}{54.67}  & 58.16    \\ \hline
\multicolumn{1}{l|}{CutPaste\&Find}        & \multicolumn{1}{c|}{\textbf{72.33}} & \multicolumn{1}{c|}{\textbf{66.67}}     & \multicolumn{1}{c|}{\textbf{89.33}}  & \textbf{76.35}    \\ \hline
\end{tabular}

}
\caption{Results on RBench. The best performances within each setting are \textbf{bolded}. $\diamondsuit$ We run these models using the OpenAI GPT-4o version. Instead of passing non-existent claims as done in their implementation, we convert all yes-no questions into affirmative statements.}

\label{tab:main2}
\end{center}

\vspace{-10pt}
\end{table}

\section{Experiment}

\subsection{Experimental Settings}

\subsubsection{Benchmark}

\paragraph{POPE}~\cite{Li-hallucination-2023}: POPE evaluates hallucinations in LVLMs using three sampling strategies: random, popular, and adversarial, each differing in negative sample construction. The random strategy selects objects absent from the image, popular samples frequent but non-existent objects, and adversarial picks commonly co-occurring but missing objects. Following~\cite{li2023evaluating}, we sampled 50 images, generating six questions per image with an equal mix of positive and negative samples (50\%). Object annotations were converted into binary Yes-or-No questions, focusing on existence hallucinations. LVLMs determine object presence, with performance measured via accuracy, precision, recall, and F1-score.

\paragraph{R-Bench}~\cite{pmlr-v235-wu24l}: The Relationship Hallucination Benchmark (R-Bench) evaluates relationship hallucinations in LVLMs. It comprises 11,651 binary questions: 7,883 image-level for global relationships and 3,768 instance-level for localized ones, using bounding boxes or masks. Evaluations on popular LVLMs reveal relationship hallucinations are more prevalent than object-level ones due to long-tail distributions and co-occurrence biases, highlighting challenges in spatial reasoning and reliance on commonsense over visual cues. R-Bench promotes improvements in mitigating relationship hallucinations via fine-grained image-text alignment.

\subsubsection{Implementation}

\paragraph{Baselines.}

We choose mainstream LVLMs as our baseline models, including mPLUG-Owl~\cite{ye2023mplug}, LLaVA~\cite{liu2023visual}, Qwen~\cite{Qwen-VL}. Regarding LVLM hallucination detection, we compare our work with Woodpecker~\cite{yin2023woodpecker} and LogicCheckGPT (LCGPT)~\cite{wu2024logical}.

\paragraph{Implementation Details.}

Our pipeline is training-free and comprises several pre-trained models apart from the LVLM to be corrected. 

\begin{itemize}[leftmargin=*]
    \item Scene Graph Parser: we use FlanT5 pretrained with FACTUAL~\cite{li-etal-2023-factual} to extract textual scene graph.
    \item Open-set object detection: we use Grounding DINO~\cite{liu2023grounding} to extract object counting information with default detection thresholds.
    \item Textual encoder, visual encoder: we use pretrained CLIP~\cite{radford2021learningtransferablevisualmodels} for extracting text and visual representation.
    \item Visual question answering: we use BLIP-2-FlanT5$_\text{XXL}$~\cite{li2023blip} as the VQA model to answer the question about: existence, attribute, and relation.
    \item Datastore: To search over this large datastore, we use FAISS \cite{johnson2017billion}, an open-source library for fast nearest neighbor retrieval in high dimensional spaces.
\end{itemize}

For both POPE and R-Bench benchmark, they are both ``Yes-or-No'' questions, instead of feeding the question and the answer from LVLMs, we use QA2Claim model~\cite{huang-etal-2023-zero} to convert all questions into claims which assume the answer is ``Yes''. For example, given a question, ``\texttt{Is there a dog in the image?}'', we convert into the specific caption ``\texttt{There is a dog in the image.}''.


\subsection{Experimental Results}


\subsubsection{POPE}

Table~\ref{tab:main} presents the performance of all methods across the different data splits. Our method consistently outperforms both baselines in all metrics and data splits, demonstrating superior robustness and generalization capabilities.

\vspace{-4pt}
\paragraph{Adversarial Split}

In the adversarial split, which challenges the models with difficult examples designed to exploit their weaknesses, our method achieves an Accuracy of 94.67\%, surpassing Woodpecker (90.67\%) and Logic (83.33\%). Notably, our method maintains a high Precision of 92.95\% and an outstanding Recall of 96.67\%, leading to the highest F1 Score of 94.77\%. This indicates that our model is not only accurate but also reliable in identifying true positives even under adversarial conditions.

\vspace{-4pt}
\paragraph{Popular Split}

For the popular split, which includes frequently occurring patterns, our method achieves an Accuracy of 93.67\%, significantly higher than Woodpecker (89.67\%) and Logic (85.00\%). The F1 Score of 93.65\% highlights our method's balanced performance in terms of Precision (93.96\%) and Recall (93.33\%), confirming its effectiveness in handling common data patterns without overfitting.

\vspace{-4pt}
\paragraph{Random Split}

In the random split, designed to reflect general, unbiased data distribution, our method achieves the highest Accuracy of 95.67\%, outperforming Woodpecker (93.33\%) and Logic (85.67\%). With a Precision of 95.36\% and Recall of 96.00\%, our method attains an exceptional F1 Score of 95.68\%, demonstrating its strong generalization capability across diverse data samples.

\subsubsection{R-Bench}

Table~\ref{tab:main2} summarizes the performance of all methods on the R-Bench benchmark, evaluated at both the image and instance levels. Our method consistently demonstrates superior performance, underscoring its robustness and adaptability across different granularity levels.

\vspace{-4pt}
\paragraph{Image Level}

At the image level, our method achieves an Accuracy of 79.00\%, significantly outperforming Woodpecker (55.00\%) and Logic (62.00\%). The Precision of 74.58\% and outstanding Recall of 88.00\% lead to the highest F1 Score of 80.73\%. These results highlight our model's effectiveness in accurately identifying relevant features within images.

\vspace{-4pt}
\paragraph{Instance Level}

For instance-level evaluation, our method achieves an Accuracy of 72.33\%, compared to Woodpecker (56.67\%) and Logic (60.67\%). The Precision of 66.67\% and remarkable Recall of 89.33\% result in an F1 Score of 76.35\%, indicating our model's strong capability to generalize well at a finer granularity, effectively capturing individual instances within images.

\begin{table}[t]

\scalebox{0.83}{
\begin{tabular}{lcccc}
\hline
\multicolumn{1}{l|}{Method}    & \multicolumn{1}{c|}{Acc}   & \multicolumn{1}{c|}{Precision} & \multicolumn{1}{c|}{Recall} & F1 Score \\ \hline

\multicolumn{5}{c}{\textit{Adversarial}     }                                                                                                 \\ \hline
\multicolumn{1}{l|}{LLaVA}     & \multicolumn{1}{c|}{50.77} & \multicolumn{1}{c|}{50.39}     & \multicolumn{1}{c|}{\textbf{99.87}}  & 66.98    \\ \hline
\multicolumn{1}{l|}{mPLUG-Owl} & \multicolumn{1}{c|}{50.67} & \multicolumn{1}{c|}{50.34}     & \multicolumn{1}{c|}{\textbf{99.33}}  & 66.82    \\ \hline
\multicolumn{1}{l|}{CutPaste\&Find}       & \multicolumn{1}{c|}{\textbf{81.26}} & \multicolumn{1}{c|}{\textbf{77.92}}     & \multicolumn{1}{c|}{87.27}  & \textbf{82.33}    \\ \hline

\multicolumn{5}{c}{\textit{Popular}        }                                                                                                  \\ \hline
\multicolumn{1}{l|}{LLaVA}     & \multicolumn{1}{c|}{52.43} & \multicolumn{1}{c|}{51.25}     & \multicolumn{1}{c|}{\textbf{99.80}}  & 66.79    \\ \hline
\multicolumn{1}{l|}{mPLUG-Owl} & \multicolumn{1}{c|}{50.63} & \multicolumn{1}{c|}{50.32}     & \multicolumn{1}{c|}{\textbf{99.27}}  & 67.72    \\ \hline
\multicolumn{1}{l|}{CutPaste\&Find}       & \multicolumn{1}{c|}{\textbf{89.83}} & \multicolumn{1}{c|}{\textbf{91.99}}     & \multicolumn{1}{c|}{87.27}  & \textbf{89.57}    \\ \hline

\multicolumn{5}{c}{\textit{Random}    }                                                                                                       \\ \hline
\multicolumn{1}{l|}{LLaVA}     & \multicolumn{1}{c|}{54.43} & \multicolumn{1}{c|}{52.32}     & \multicolumn{1}{c|}{\textbf{99.80}}  & 68.65    \\ \hline
\multicolumn{1}{l|}{mPLUG-Owl} & \multicolumn{1}{c|}{53.30} & \multicolumn{1}{c|}{51.71}     & \multicolumn{1}{c|}{\textbf{99.53}}  & 68.06    \\ \hline
\multicolumn{1}{l|}{CutPaste\&Find}       & \multicolumn{1}{c|}{\textbf{86.80}} & \multicolumn{1}{c|}{\textbf{86.46}}     & \multicolumn{1}{c|}{87.27}  & \textbf{86.86}    \\ \hline

\end{tabular}
}

\caption{Results on POPE (Full version). The best performances within each setting are \textbf{bolded}.}

\label{tab:main3}

\vspace{-5pt}
\end{table}

\section{Ablation study}

To further validate our approach, we conducted an extensive evaluation using the full POPE dataset (9000 samples), and the results are presented in Table 2. Our method is compared against two additional state-of-the-art models: LLaVA and mPLUG-Owl.

\vspace{-4pt}
\paragraph{Random Split}

Our method achieves an Accuracy of 86.80\%, significantly outperforming LLaVA (54.43\%) and mPLUG-Owl (53.30\%). The Precision of 86.46\% and Recall of 87.27\% lead to the highest F1 Score of 86.86\%, demonstrating strong generalization across randomly distributed samples.

\vspace{-4pt}
\paragraph{Popular Split}

For the popular split, our method again achieves superior performance with an Accuracy of 89.83\%, compared to LLaVA (52.43\%) and mPLUG-Owl (50.63\%). The F1 Score of 89.57\% highlights our model’s ability to capture frequently occurring patterns while maintaining high precision (91.99\%) and recall (87.27\%).

\vspace{-4pt}
\paragraph{Adversarial Split}

In the adversarial setting, which is particularly challenging, our method maintains an Accuracy of 81.26\%, a substantial improvement over LLaVA (50.77\%) and mPLUG-Owl (50.67\%). The F1 Score of 82.33\% further confirms its robustness, balancing Precision (77.92\%) and Recall (87.27\%).

\subsection{Experiment on full POPE}

To highlight the necessity of our scaling factor, we analyze similarity scores in the visual-aid knowledge base. As shown in Table~\ref{tab:main3}, ground-truth validation images often receive suboptimal similarity scores despite correctly representing the queried object. For example, a wooden medicine cabinet scores 0.558, with alternative images varying from 0.383 to 0.805. Similarly, white fridge and train on track validation images score 0.922 and 0.655, while some alternatives score even higher (e.g., 0.823 for white fridge and 0.845 for train on track), despite contextual differences. These inconsistencies suggest that raw similarity scores may not fully capture object fidelity, leading to misinterpretations in hallucination detection. Our scaling factor mitigates this by incorporating prior knowledge from the datastore, aligning scores with realistic expectations and improving hallucination sensitivity.

\subsection{VQA scoring visualization }

To illustrate the necessity of our scaling factor, we analyze similarity scores in the visual-aid knowledge base. As shown in Figure~\ref{fig:ab2}, ground-truth validation images often receive lower-than-ideal similarity scores despite correctly representing the queried object. For example, a wooden medicine cabinet scores 0.558, while alternative images range from 0.383 to 0.805. Similarly, white fridge and train on track validation images score 0.922 and 0.655, yet some alternative images score even higher (e.g., 0.823 and 0.845, respectively), despite contextual differences. These inconsistencies indicate that raw similarity scores do not fully capture object fidelity, leading to misinterpretations in hallucination detection. Our scaling factor mitigates this by incorporating prior knowledge from the datastore, aligning scores with realistic expectations and improving sensitivity to hallucinations.

\subsection{Visualization}

\subsubsection{Open-set Object detector based comparison}





The qualitative examples in Figure~\ref{fig:ab1a} demonstrate our model’s effectiveness in detecting hallucinations, particularly where Woodpecker generates incorrect or misleading responses. In the first case, when asked about a sports ball, Woodpecker incorrectly states that none is present, while our model correctly identifies it with an existence score of 1.0, avoiding the hallucination of a missing entity. Similarly, for a dining table, Woodpecker misidentifies a generic table, whereas our model assigns an existence score of 0.0, accurately detecting the hallucination.
In the black leather couch example, Woodpecker detects a couch but fails to confirm its leather attribute, creating ambiguity. Our model explicitly identifies the leather attribute with high confidence (0.9228) and captures relational understanding by confirming the woman’s sitting position with a score of 0.9431, which Woodpecker lacks.
Lastly, for the smiling baby, Woodpecker falsely asserts the baby is smiling, while our model assigns a lower smiling attribute score (0.3378), reflecting uncertainty and mitigating hallucinated affirmations.
These examples highlight our model’s robustness in leveraging existence, attribute confidence scores, and relational reasoning to mitigate hallucinations, outperforming Woodpecker in multimodal verification.

\subsubsection{Failed cases}

The examples in Figure~\ref{fig:ab1b} illustrate failure cases where our model struggles with complex language structures, particularly in fine-grained attributes and relational constraints. In the first case, the question asks whether there are trees without lights outside a strip mall. While our model correctly detects trees and the strip mall (existence score: 1.0) and their spatial relationship (0.8866), it fails to distinguish decorated from undecorated trees, revealing a limitation in attribute-based reasoning for multi-condition queries.
In the second case, the question asks whether two women are playing table tennis against each other. The model detects women and table tennis (existence score: 1.0) and identifies the women, play, table tennis relation (0.6176). However, it fails to capture the against each other condition, requiring a deeper understanding of competitive interactions rather than simple activity recognition.
These cases highlight our model’s difficulty with compositional language, negation, and multi-entity interactions. To overcome these limitations, improvements in linguistic grounding and multimodal reasoning are needed to better handle complex queries.
\section{Conclusion}

In this work, we introduced CutPaste\&Find, a novel, training-free framework for detecting hallucinations in LVLM-generated descriptions. Unlike existing methods that depend on costly LVLM-based inference, our approach utilizes off-the-shelf visual and linguistic tools, making it both efficient and scalable. By incorporating a visual-aid knowledge base and a scaling factor for improved similarity alignment, our method enables robust hallucination detection with minimal computational overhead. Experimental results on benchmark datasets demonstrate that CutPaste\&Find is both effective and resource-efficient, offering a practical solution for enhancing the reliability of LVLMs.
\section*{Limitations}

The proposed method relies heavily on the Visual Genome dataset for constructing its visual-aid knowledge base, which limits its generalizability to domains lacking high-quality, well-annotated datasets. Expanding to domain-specific or low-resource scenarios may require significant manual effort to curate new knowledge bases. Additionally, while the approach effectively verifies object existence, attributes, and relations, it struggles with fine-grained semantic understanding, particularly in handling compositional queries, negations, and multi-entity interactions. For instance, it may fail to distinguish between "two women playing against each other" and simply "playing table tennis," highlighting a limited ability to process intricate logical dependencies in multimodal reasoning.
\bibliography{custom}

\appendix
\appendix
\onecolumn

\begin{figure}[t]
    \centering
    \includegraphics[width=0.8\textwidth]{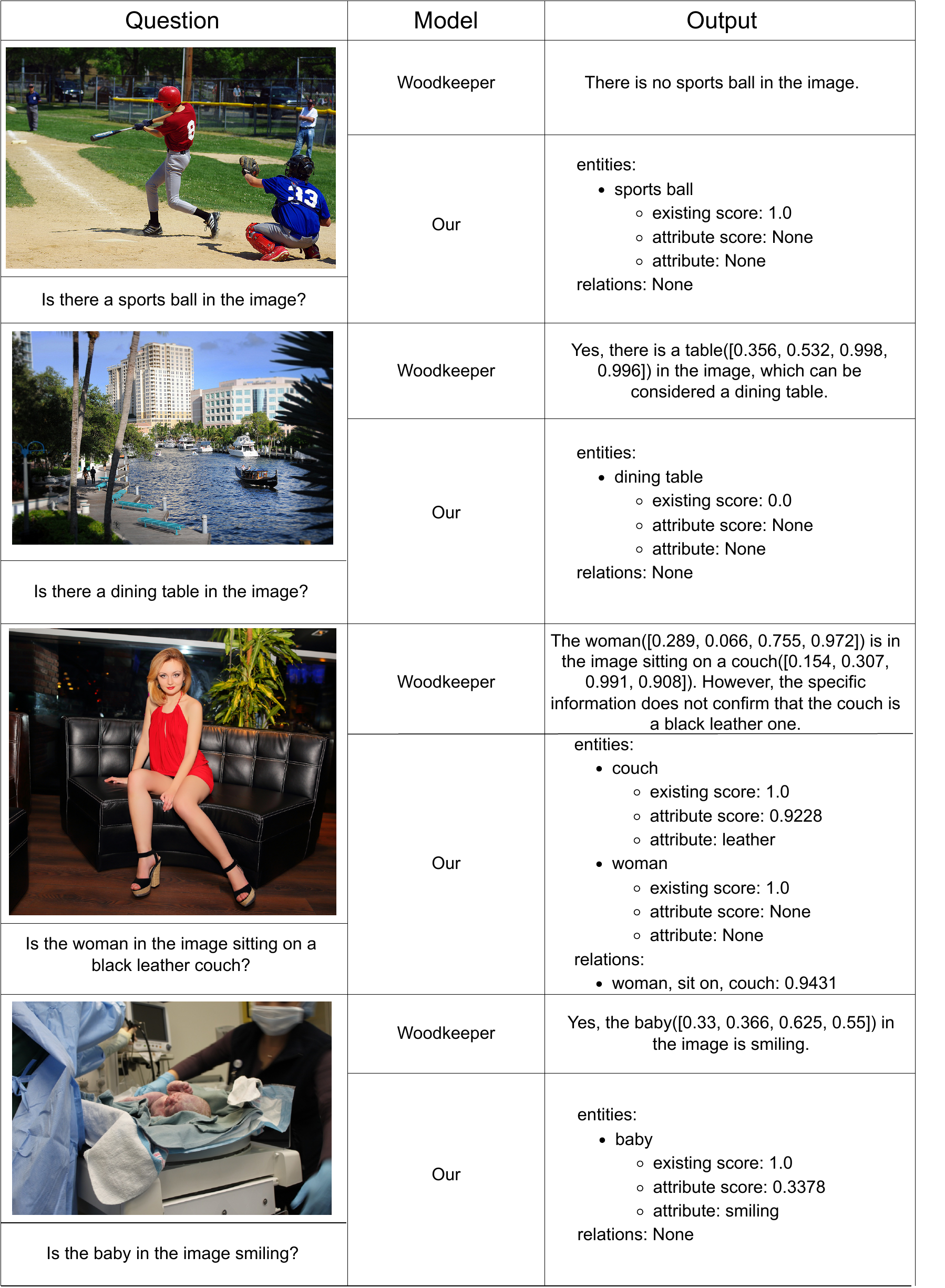}
    \caption{Examples of inference between Woodpecker and our CutPaste\&Find.}
    \label{fig:ab1a}
\end{figure}

\begin{figure}[t]
    \centering
    \includegraphics[width=0.8\textwidth]{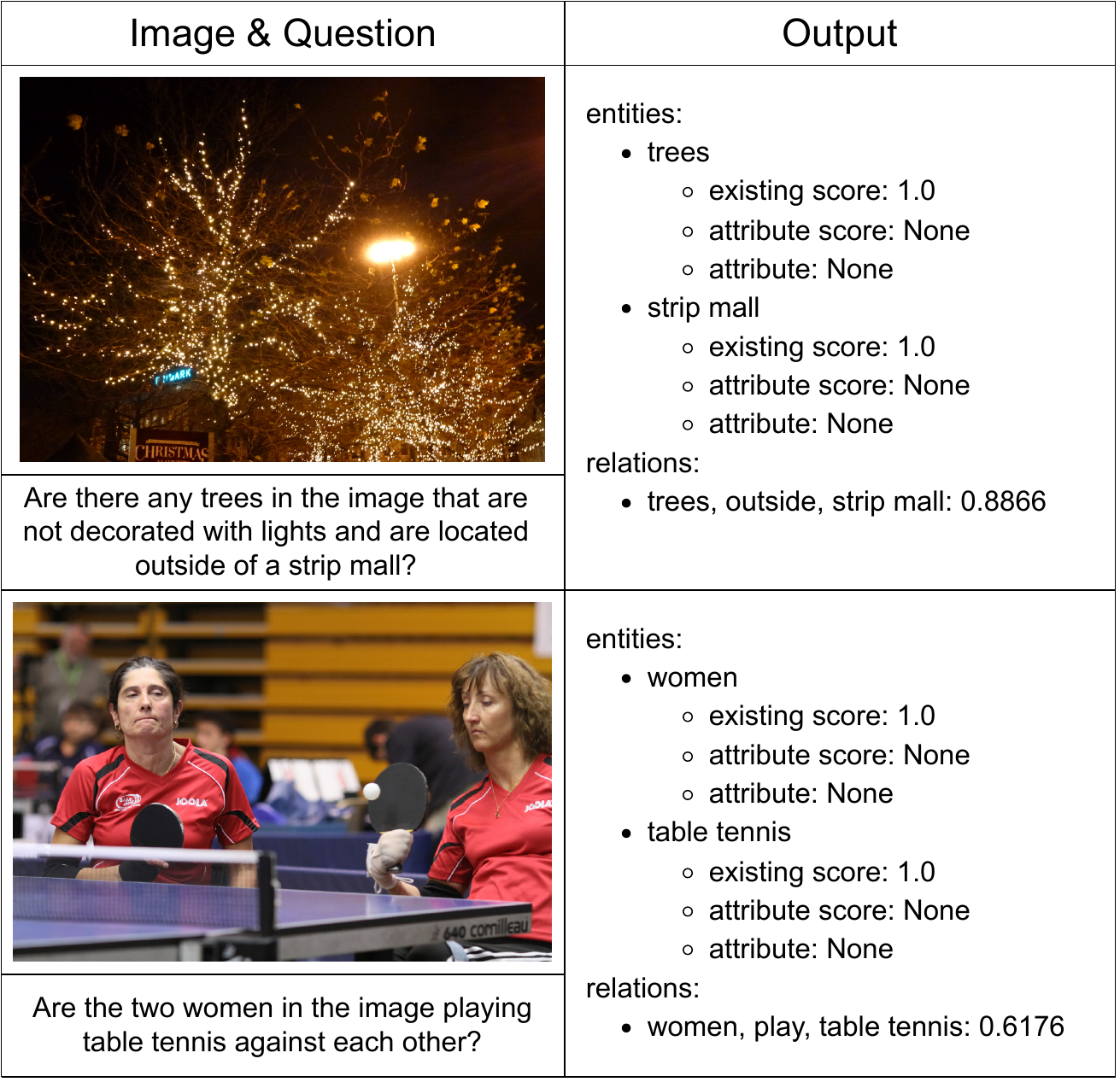}
    \caption{Failed examples of our CutPaste\&Find.}
    \label{fig:ab1b}
\end{figure}

\begin{figure}[t]
    \centering
    \includegraphics[width=0.99\textwidth]{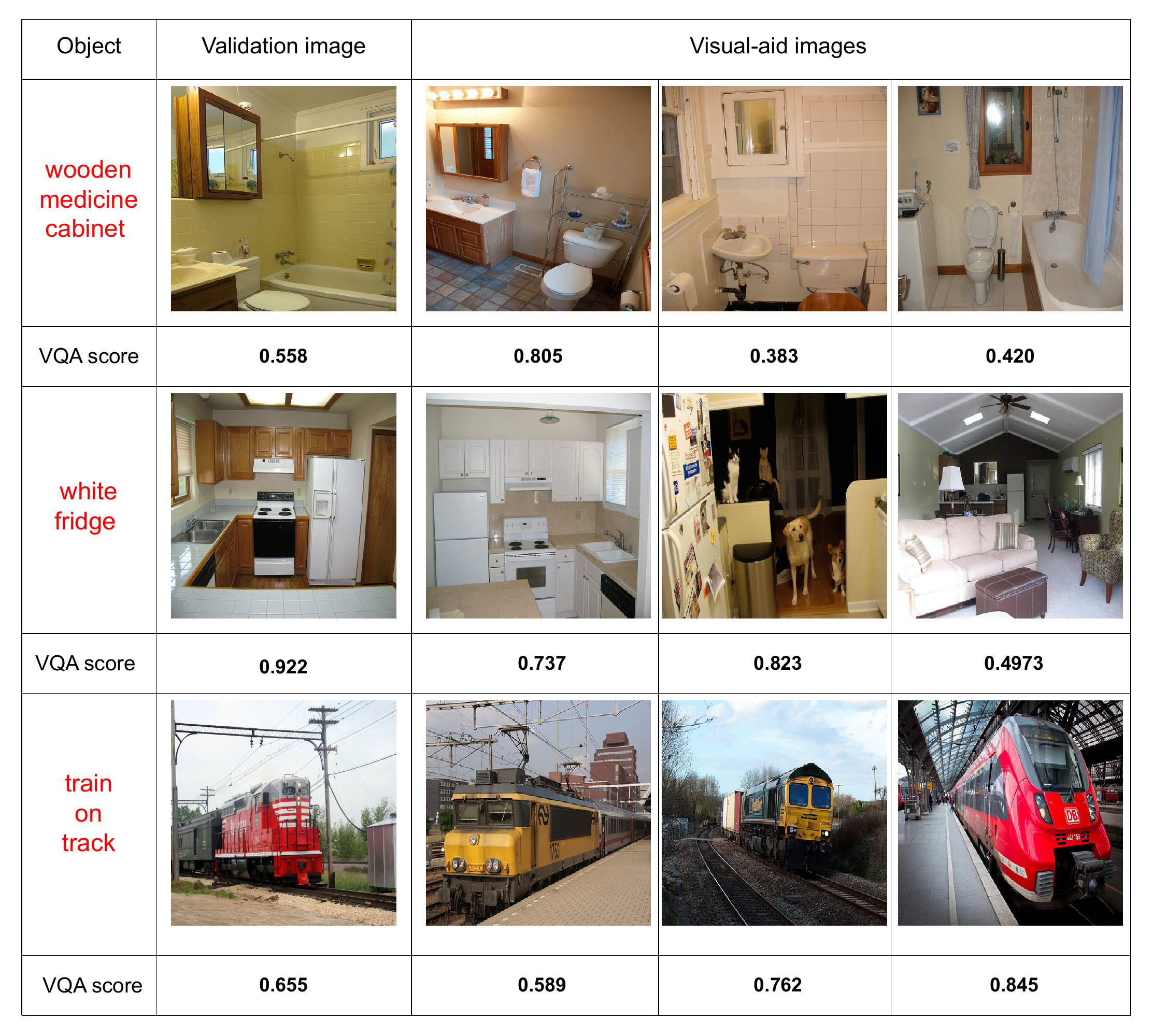}
    \caption{Visualization of VQA module on Knowledge base images and test images.}
    \label{fig:ab2}
\end{figure}


\label{sec:appendix}

\end{document}